\documentclass[conference]{IEEEtran}
\IEEEoverridecommandlockouts
\usepackage{cite}
\usepackage{float}  
\usepackage{graphicx}
\usepackage{subcaption}
\usepackage{ragged2e}
\usepackage{blindtext}
\usepackage{url}
\usepackage{subcaption} 
\usepackage{float} 
\usepackage{balance}
\usepackage{stfloats} 
\usepackage{amsmath,amssymb,amsfonts}
\usepackage{booktabs}
\usepackage{algorithmic}
\usepackage{graphicx}
\usepackage{textcomp}
\usepackage{xcolor}
\def\BibTeX{{\rm B\kern-.05em{\sc i\kern-.025em b}\kern-.08em
    T\kern-.1667em\lower.7ex\hbox{E}\kern-.125emX}}
\begin{document}
\makeatletter
\newcommand{\linebreakand}{%
  \end{@IEEEauthorhalign}
  \hfill\mbox{}\par
  \mbox{}\hfill\begin{@IEEEauthorhalign}
}
\makeatother
\title{LLM-Driven AutoML for Cross-Lingual Handwritten OCR: Closed-Loop Neural Architecture Search with GPT-5, GPT-4o, and Claude Sonnet 4}

\author{
\IEEEauthorblockN{Mobina Kashaniyan}
\IEEEauthorblockA{
\textit{Department of Computer Engineering} \\
\textit{Iran University of Science and Technology} \\
Tehran, Iran \\
}
\and
\IEEEauthorblockN{Amirhossein Ghassemi}
\IEEEauthorblockA{
\textit{Department of Computer Engineering} \\
\textit{Iran University of Science and Technology} \\
Tehran, Iran \\
}
\linebreakand
\IEEEauthorblockN{Nasser Mozayani}
\IEEEauthorblockA{
\textit{Department of Computer Engineering} \\
\textit{Iran University of Science and Technology} \\
Tehran, Iran \\
}
}

\maketitle


\maketitle

\begin{center}
\footnotesize
\copyright~2025 IEEE. Personal use of this material is permitted.
Permission from IEEE must be obtained for all other uses, in any current
or future media, including reprinting/republishing this material for
advertising or promotional purposes, creating new collective works,
for resale or redistribution to servers or lists, or reuse of any
copyrighted component of this work in other works.

\vspace{0.5em}

This is the accepted version of the paper published in the
2025 15th International Conference on Computer and Knowledge Engineering (ICCKE),
IEEE, 2025. DOI: 10.1109/ICCKE68588.2025.11273810.
\end{center}

\vspace{0.5em}

\begin{abstract}
Handwritten text recognition across diverse scripts presents an enduring challenge in machine learning, as each language and writing system introduces unique visual complexities and demands. Traditional approaches have depended on expert-guided model design and extensive preprocessing, which make it difficult to scale and adapt to new scripts efficiently. In this work, we introduce a pipeline that is fully automatic and cross lingual, using large language models, GPT 5, GPT 4o and Claude Sonnate 4, to independently generate, evaluate, and refine neural network architectures for handwritten optical character recognition. This process requires no manual intervention, domain specific preprocessing, or human selection of models, resulting in a complete end to end automated system.

We apply this approach to Arabic, English, and Persian scripts, each representing distinct character shapes and writing traditions, and conduct thirty independent trials for every language. The pipeline consistently discovers efficient models with high test accuracy, achieving average scores above ninety three percent, while also maintaining inference speeds that meet the needs of real time applications. Notably, the system is able to automatically explore a wide range of neural architectures and adaptively select designs that fit the unique requirements of each script, without any explicit guidance from human experts. These results show that large language models can move beyond language processing and act as independent designers for machine learning systems. This enables a scalable, script agnostic, and fully automatic solution for multilingual handwritten text recognition, opening the door to rapid and adaptable deployment of OCR technology across many languages and domains.
\end{abstract}
\begin{IEEEkeywords}
large language models, neural architecture search, handwritten text recognition, multilingual OCR, automation, model discovery
\end{IEEEkeywords}
\IEEEpubidadjcol
\section{Introduction}  
Converting handwritten text into digital formats is increasingly essential for professionals such as editors, students, archivists, and researchers. Despite significant advances in artificial intelligence, accurately recognizing handwritten characters remains a complex challenge across diverse scripts such as Arabic, English, and Persian. These languages present unique structural characteristics, including intricate stroke patterns, diacritics, and high variability in writing styles among individuals. Traditional Optical Character Recognition (OCR) approaches typically rely on manually designed neural architectures, such as convolutional or transformer-based networks. While these handcrafted models have achieved promising results, they often demand extensive expert intervention to iteratively adjust network layers, tune hyperparameters, or redesign model components. This process can be both time-consuming and prone to inconsistent outcomes. Consequently, prior work has often prioritized elaborate preprocessing techniques rather than addressing architectural improvements directly. Recent developments in large language models (LLMs), particularly GPT-5, GPT-4o, and Claude Sonnet 4, have opened new frontiers for automated neural network design. These models demonstrate advanced reasoning and code-generation capabilities, enabling them to serve as intelligent agents for architectural discovery. Motivated by this emerging capability, we present a fully automated pipeline for multilingual handwritten text recognition in Arabic, English, and Persian scripts, where GPT-5, GPT-4o, and Claude Sonnet 4 are employed independently as AI architects responsible for proposing, refining, and optimizing deep learning models.  

Our pipeline begins with raw handwritten image data and minimal augmentation. Dataset-specific metadata including the number of classes and image dimensions is used to prompt GPT-5, GPT-4o, and Claude Sonnet 4 to propose candidate model architectures in structured JSON format. These architectures incorporate a variety of components including convolutional layers, pooling operations, normalization, dropout, and optional transformer-based modules. Each proposed model is then trained and evaluated without human intervention. Following each trial, performance metrics such as training accuracy, validation accuracy, and test accuracy are fed back into the same LLM that generated the architecture. This creates a closed feedback loop that allows each LLM to refine its subsequent proposals based on previous outcomes. This iterative architecture optimization process eliminates the need for manual hyperparameter tuning and significantly accelerates the model development cycle. While recent research has leveraged LLMs for direct image or text classification, their role as generative agents for automated model discovery remains largely unexplored. To the best of our knowledge, our approach is the first to apply GPT-5, GPT-4o, and Claude Sonnet 4 in closed-loop pipelines for architecture generation in the domain of cross-lingual handwritten text recognition.  

The key contributions of this work are summarized below:  

\begin{itemize}  
    \item \textbf{Cross-script Capability:} We introduce a single, unified framework capable of recognizing multiple handwritten scripts, including Arabic, English, and Persian. Each language benefits from tailored preprocessing and encoding strategies to accommodate its specific structural and visual complexities.  

    \item \textbf{LLM-driven Model Discovery:} GPT-5, GPT-4o, and Claude Sonnet 4 are employed as standalone architecture generators. By translating dataset-specific insights into executable model configurations, our method entirely removes the need for manual architecture design.  

    \item \textbf{Automated Iterative Refinement:} The system operates in a closed feedback loop in which each LLM receives trial-by-trial results and uses them to refine future architecture proposals. This dynamic process enables convergence toward more effective solutions over successive iterations.  

    \item \textbf{Transparency and Reproducibility:} Every experimental trial is rigorously documented, with architecture configurations, training logs, and performance metrics saved in structured formats such as JSON and CSV. This design ensures full reproducibility and facilitates meaningful comparisons across model variations and LLM strategies.  
\end{itemize}  

By transforming GPT-5, GPT-4o, and Claude Sonnet 4 into autonomous architecture designers, our framework significantly simplifies OCR model development and advances the capabilities of automated machine learning for handwriting recognition. The rest of this paper is organized as follows: Section~2 discusses related work on multilingual handwritten OCR and recent trends in automated model generation. Section~3 details our methodology, including architecture prompting, model generation, training procedures, and the feedback loop. Section~4 presents experimental results across all three languages and provides an analysis of performance, parameter efficiency, and architectural evolution. Finally, Section~5 concludes the paper and outlines directions for future research.

\section{Related Work}

Recent advances in handwritten text recognition (HTR) have leveraged deep learning and neural architecture search (NAS), particularly for complex scripts like Arabic and Persian. Qalam~\cite{b1} integrates a SwinV2 vision encoder with a RoBERTa-style decoder to handle both printed and handwritten Arabic text, highlighting the power of transformers for cursive, context-sensitive features. For Persian and Arabic,~\cite{b2} uses an ensemble of feedforward networks optimized with particle swarm and League Championship algorithms, while~\cite{b10} applies DenseNet and Xception with augmentation,~\cite{b11} introduces Bina, a CNN–BiRNN trained on synthetic data, and~\cite{b12} develops a segmentation-free CNN–RNN–CTC model for offline word recognition. Broader reviews~\cite{b5,b13} categorize HTR methods, survey datasets, and outline challenges such as style variability, limited data, and low-resource language coverage. Parallel to OCR progress, large language models have been explored for NAS. GPT-NAS~\cite{b3} combines a pretrained GPT with a genetic algorithm for architecture search on NAS-Bench-101, achieving strong results on CIFAR and ImageNet, while GENIUS~\cite{b4} prompts GPT-4 to iteratively generate and refine models, rivaling conventional NAS with minimal domain expertise. While prior work addresses OCR for complex scripts and LLM-driven NAS separately, our approach unifies them by employing GPT-5, GPT-4o, and Claude Sonnet 4 in a closed-loop, multilingual OCR pipeline that automates both model generation and refinement across Arabic, English, and Persian.

\section{The Proposed Approach}
\begin{figure}
    \centering
    \includegraphics[width=0.7\linewidth]{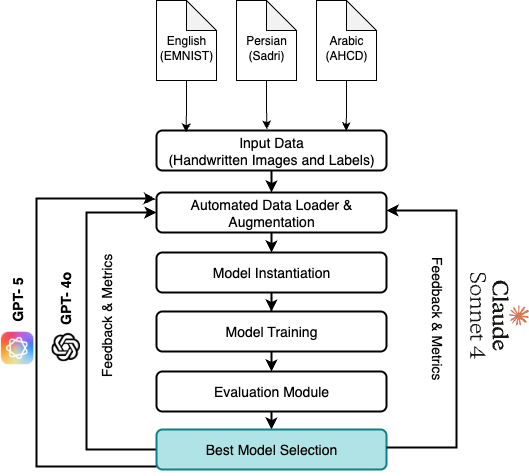}
    \caption{The fully automated LLM-driven cross-lingual OCR architecture search pipeline. Evaluation results are fed back to the LLM, which iteratively refines model designs until the best architecture is discovered for each script.}
    \label{fig:enter-label}
\end{figure}

Our end to end automated machine learning (AutoML) pipeline for Optical Character Recognition (OCR) revolves around an integrated and iterative loop comprising four core components: (1) data collection and augmentation, (2) large language model (LLM) driven neural architecture proposal, (3) automated model training and evaluation, and (4) continuous performance feedback for refining subsequent model architectures. By systematically repeating this cycle across numerous trials, our pipeline autonomously identifies and leverages optimal architectural patterns and hyperparameters for each targeted handwriting script, including Arabic, English, and Persian, without manual intervention. Figure \ref{fig:enter-label} illustrates the architecture of the proposed framework.
\subsection{Data Collection and Augmentation}
We utilize the EMNIST dataset~\cite{b7} for English, the SADRI dataset~\cite{b8} for Persian, and the AHCD dataset~\cite{b9} for Arabic handwriting recognition. For each script, we implement specialized data loaders to collect raw handwritten images along with their corresponding labels. These loaders reshape the data into standardized tensor formats and partition datasets into training, validation, and test subsets using stratified sampling to maintain class distributions, dedicating 10\% of the data to validation. To enhance the model's generalization capability while ensuring computational efficiency, we apply a lightweight data augmentation strategy and employ Keras’s ImageDataGenerator to introduce random rotations, horizontal and vertical shifts, and zoom operations directly during training. This method enriches the training set's diversity without imposing significant computational overhead.
\subsection{LLM Driven Neural Architecture Proposal}
At the beginning of each trial, we prepare a structured prompt containing essential dataset details, such as the number of classes and image dimensions, and, when available, performance metrics from the previous iteration (training, validation, and test accuracies). We provide this information to GPT-5, GPT-4o and Claude Sonnet 4, which responds with a detailed JSON specification. The specification includes the sequential structure of neural layers, such as Conv2D, BatchNorm, MaxPooling, Dropout, Flatten, Dense, and optionally, hybrid patch embedding and transformer blocks. It also specifies hyperparameters like the optimizer choice, learning rate, batch size, number of epochs, early stopping patience, and evaluation metrics. Utilizing JSON specifications ensures deterministic parsing, allowing seamless and automated integration into the subsequent model construction step.
\subsection{Model Construction and Training}
We parse the architecture specifications generated by GPT-5, GPT-4o and Claude Sonnet 4 returned in JSON format and automatically convert them into complete neural network models using Keras. Each specification defines a sequence of layers, including convolutional blocks, batch normalization, pooling, dropout, flattening, and dense layers terminating in a softmax classifier for multi-class prediction. The JSON format also contains training hyperparameters such as learning rate, optimizer type, batch size, and early stopping criteria. Our parser accommodates both conventional CNN-based structures and more advanced hybrid architectures. When specified, the system constructs models incorporating Vision Transformer components through custom implementations of PatchEmbedding and TransformerBlock layers. These modules support projection-based patch extraction and self-attention over spatial tokens, enabling exploration of token-mixing representations beyond spatial convolutions. Each model is compiled with categorical cross-entropy loss and optimized using either the Adam or SGD optimizer, as indicated in the LLM-generated specification. 

\subsection{Evaluation and Metric Collection}

After training concludes, we rigorously evaluate each model on its corresponding test set to determine final test accuracy. We also record the highest training and validation accuracies observed during the optimization process. Inference latency is measured through repeated forward passes using single sample inputs to simulate real time prediction performance. Additionally, we document the total number of trainable parameters as an indicator of model complexity. All collected metrics along with the full JSON model specification are systematically logged in a centralized CSV file. This ensures transparency, facilitates cross trial comparisons, and supports reproducibility in all downstream analyses.

\subsection{Continuous Self-Improvement Feedback Loop}

Following each training trial, we compile a structured summary containing key performance metrics training accuracy, validation accuracy, and test accuracy into a concise JSON object. This summary is passed as feedback into the subsequent LLM prompt, enabling the model to evaluate the effectiveness of its prior architectural decisions. The language model uses this information to adjust architectural elements such as layer types, depth, width, and hyperparameter settings in future proposals. This feedback loop operates automatically across multiple trials, allowing the LLM to iteratively refine its understanding of which configurations yield better results for each handwriting script. As a result, the system converges toward architectures that balance predictive performance, inference efficiency, and model complexity. 
\section{Evaluation}
\label{sec:evaluation}
\begin{figure*}[t]
    \centering
    \includegraphics[width=.32\linewidth]{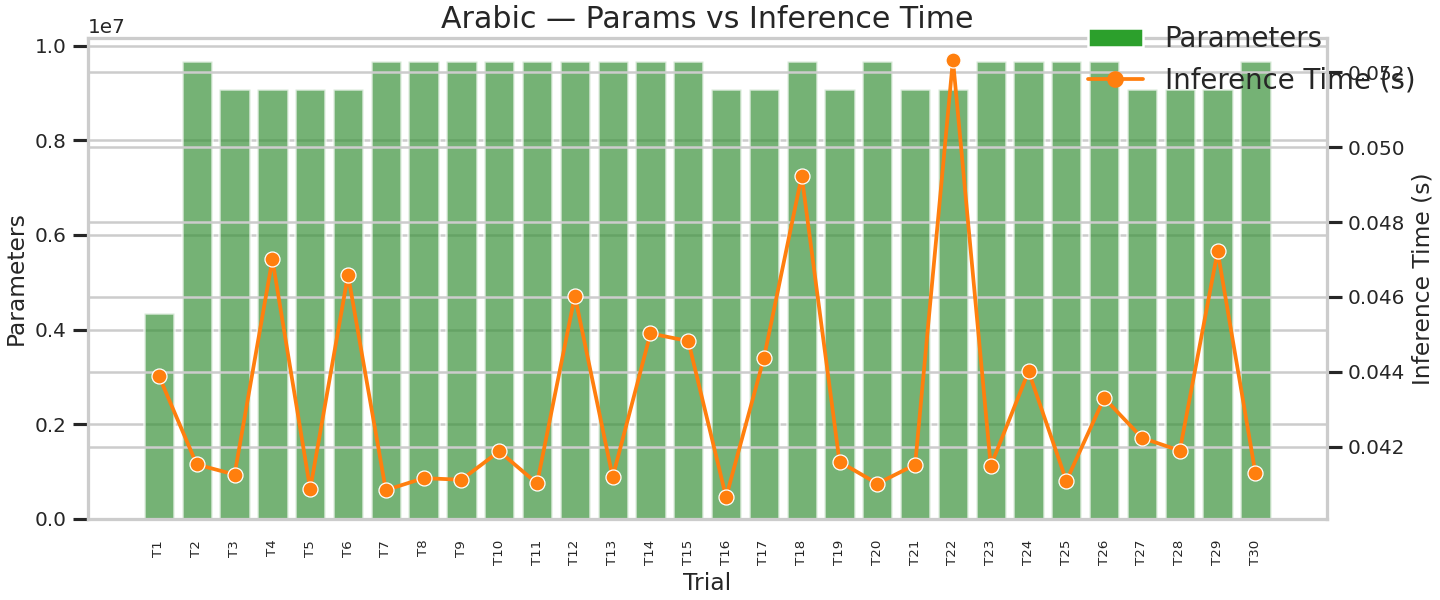}\hfill
    \includegraphics[width=.32\linewidth]{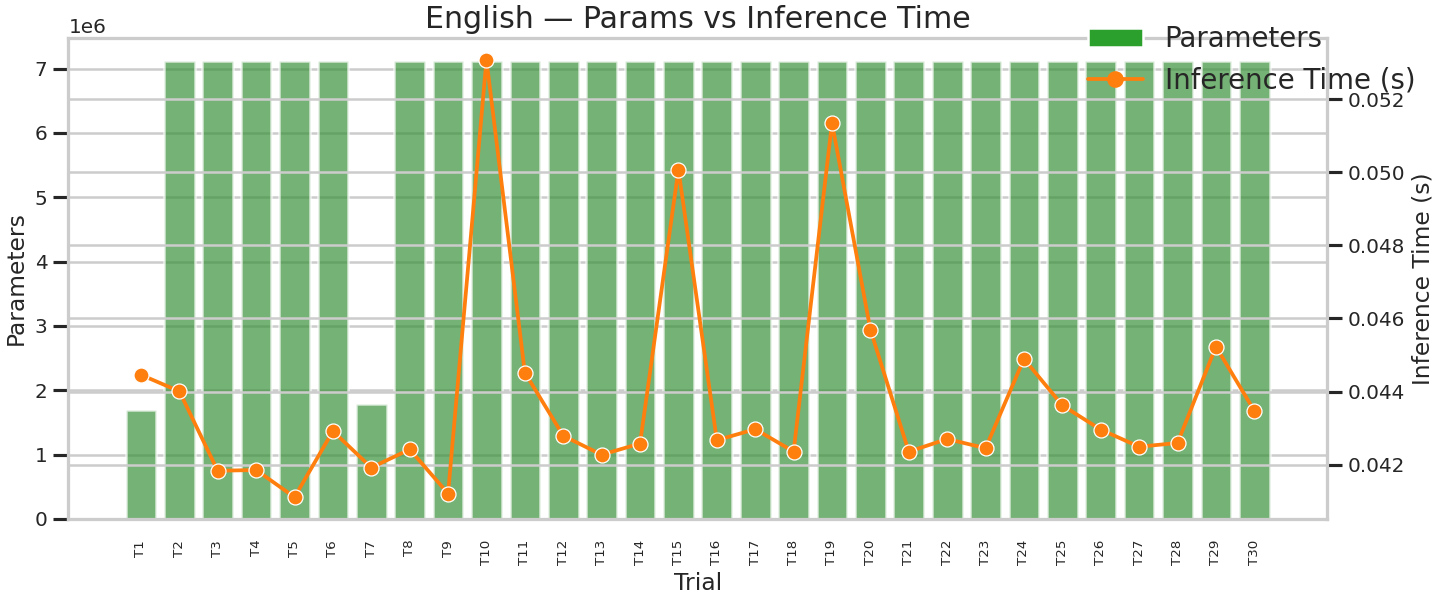}\hfill
    \includegraphics[width=.32\linewidth]{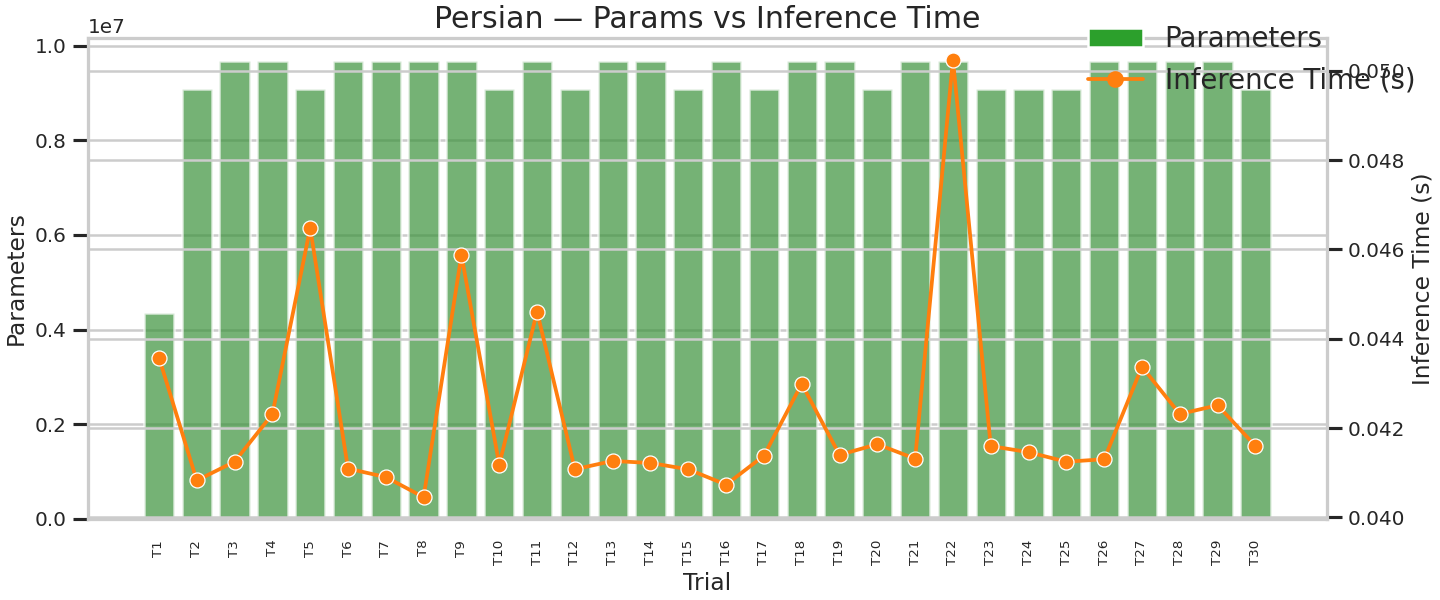}
    \caption{Claude Sonnet 4 parameter count and per sample inference time across thirty trials for Arabic, English, and Persian. Latency remains stable near forty one milliseconds despite large variation in model size, which suggests that runtime cost is driven more by architectural depth than by parameter count.}
    \label{fig:claude-paramlat}
\end{figure*}

\begin{figure*}[t]
    \centering
    \includegraphics[width=.32\linewidth]{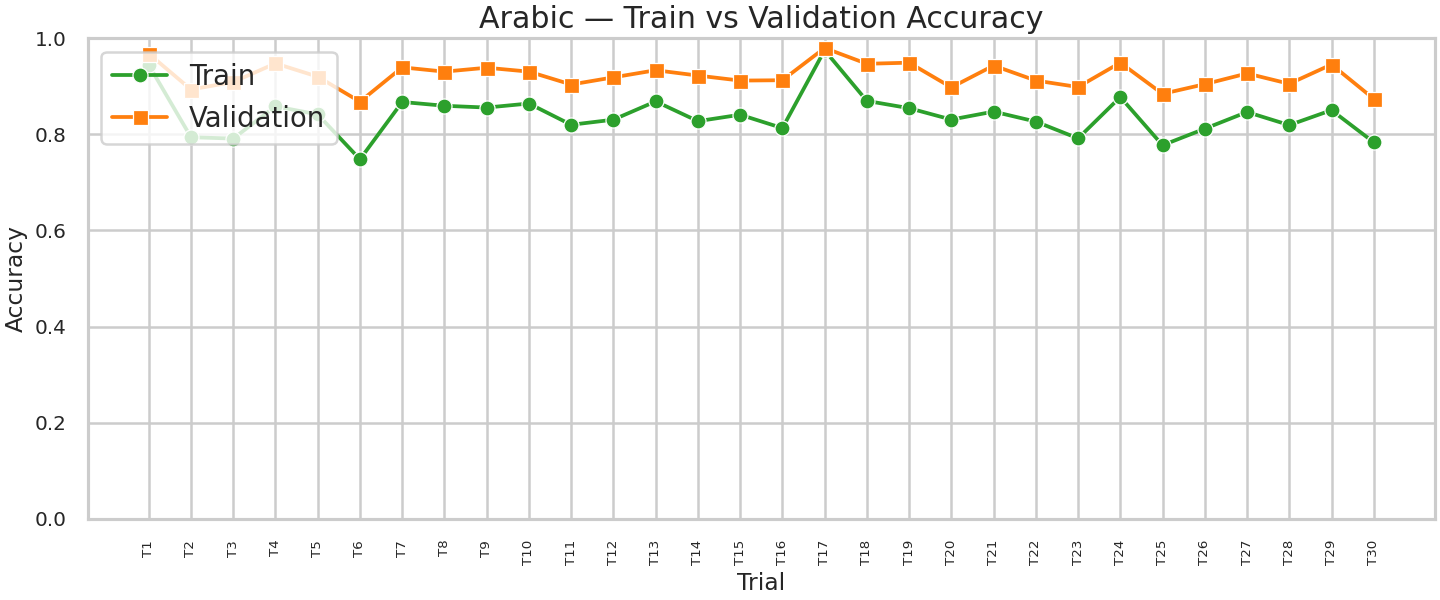}\hfill
    \includegraphics[width=.32\linewidth]{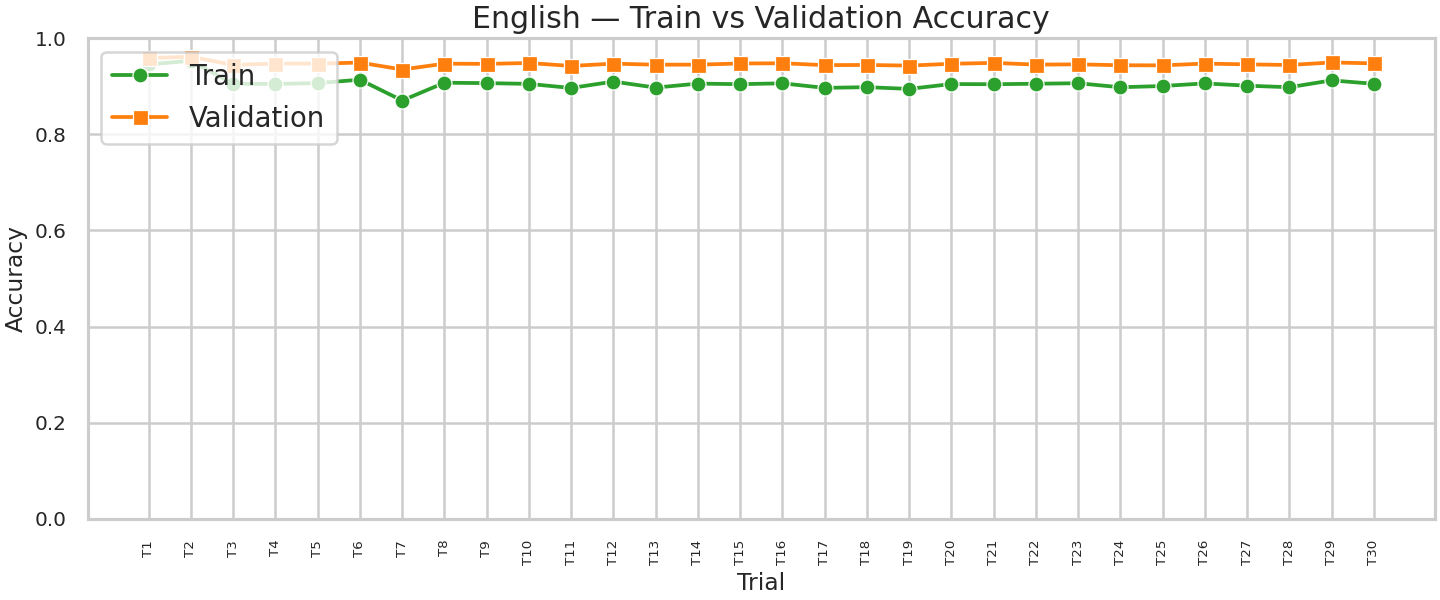}\hfill
    \includegraphics[width=.32\linewidth]{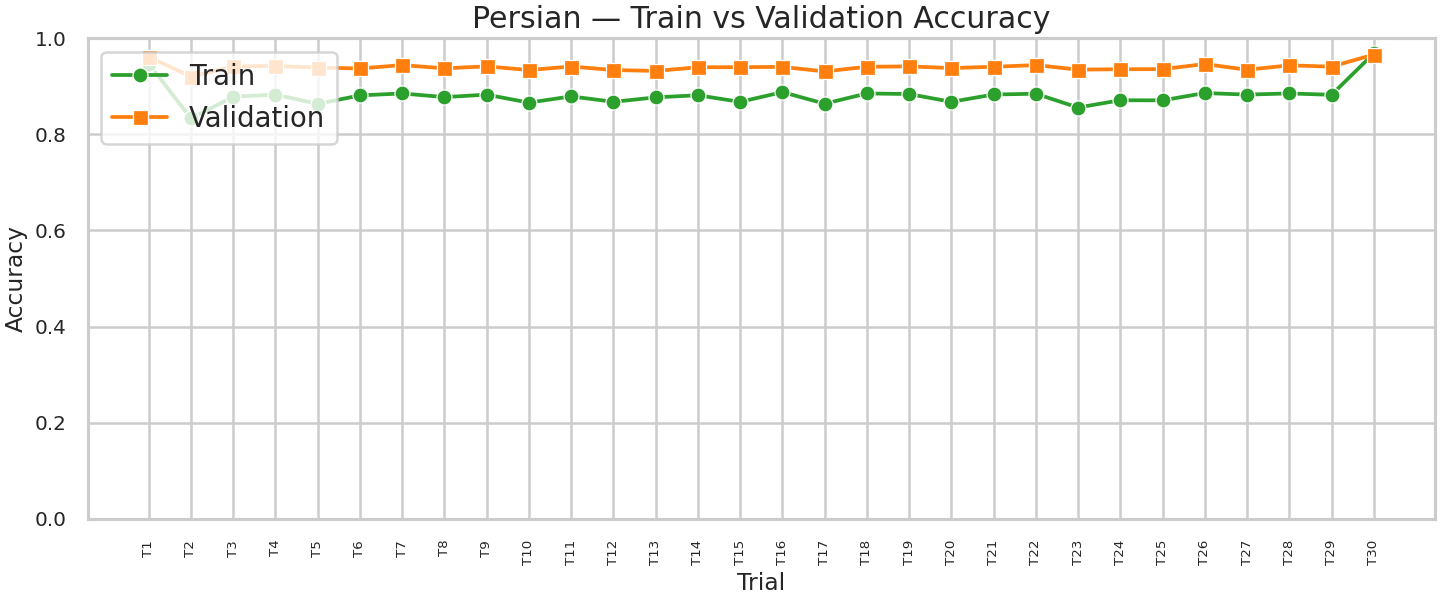}
    \caption{Claude Sonnet 4 training and validation accuracy across thirty generated models for Arabic, English, and Persian. Validation closely follows training with differences typically within one to two percentage points, which indicates strong generalization.}
    \label{fig:claude-trainval}
\end{figure*}

\begin{figure}[t]
    \centering
    \includegraphics[width=0.5\columnwidth]{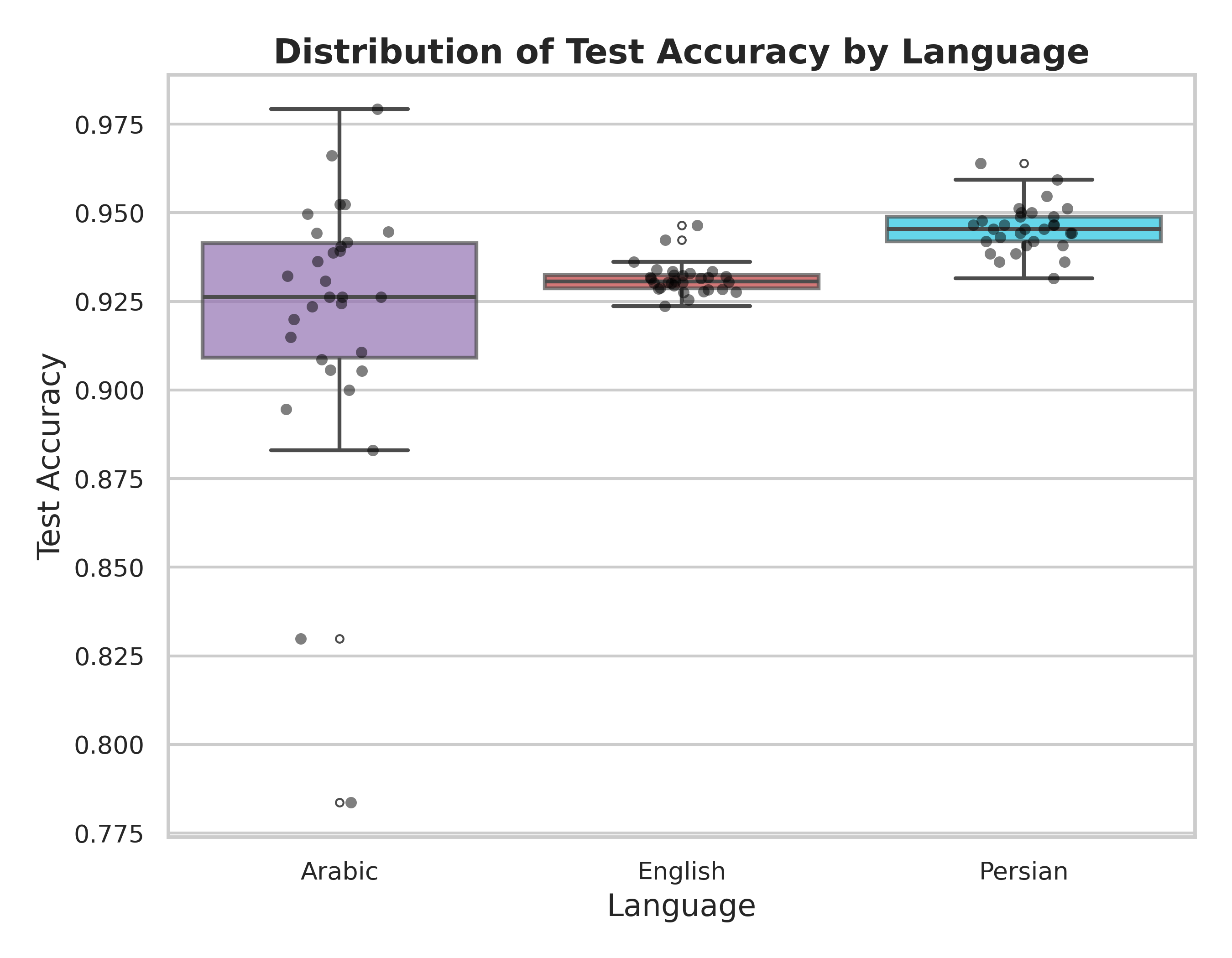}
    \caption{Claude Sonnet 4 distribution of test accuracy across thirty trials per script for Arabic, English, and Persian. Persian shows the highest median, Arabic exhibits wider spread, and English is stable.}
    \label{fig:claude-boxplot}
\end{figure}
\begin{figure*}[t]
    \centering
    \includegraphics[width=.30\linewidth]{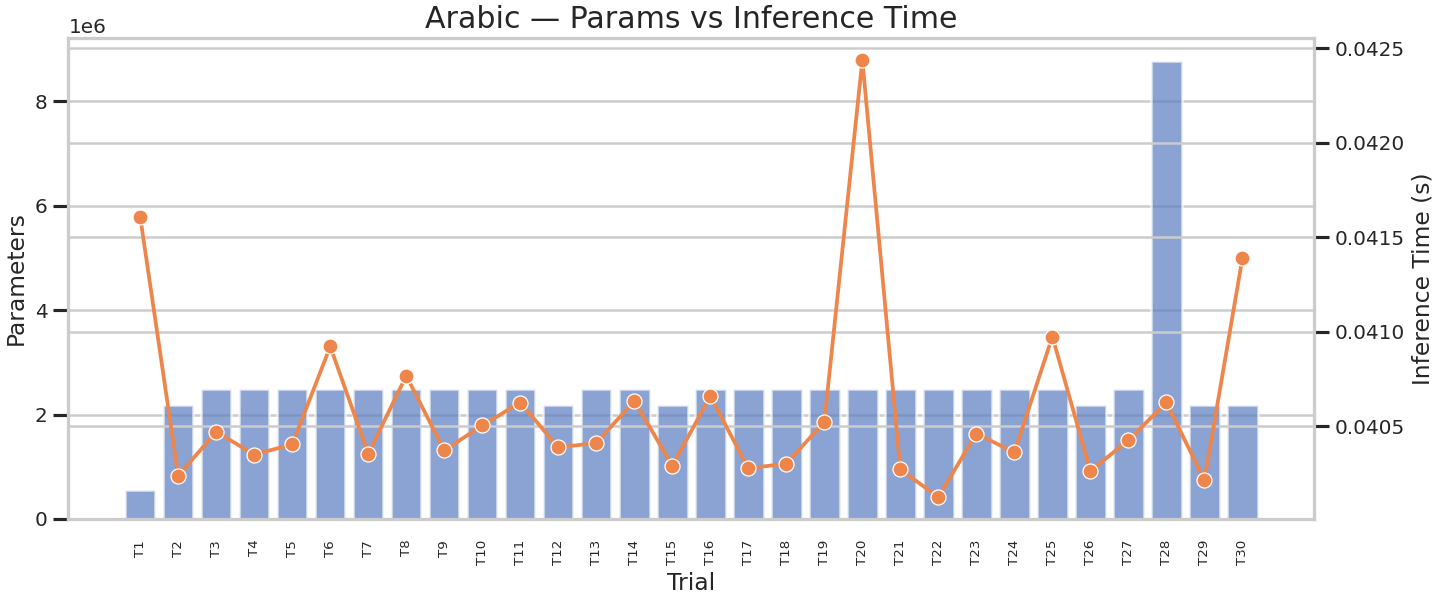}\hfill
    \includegraphics[width=.30\linewidth]{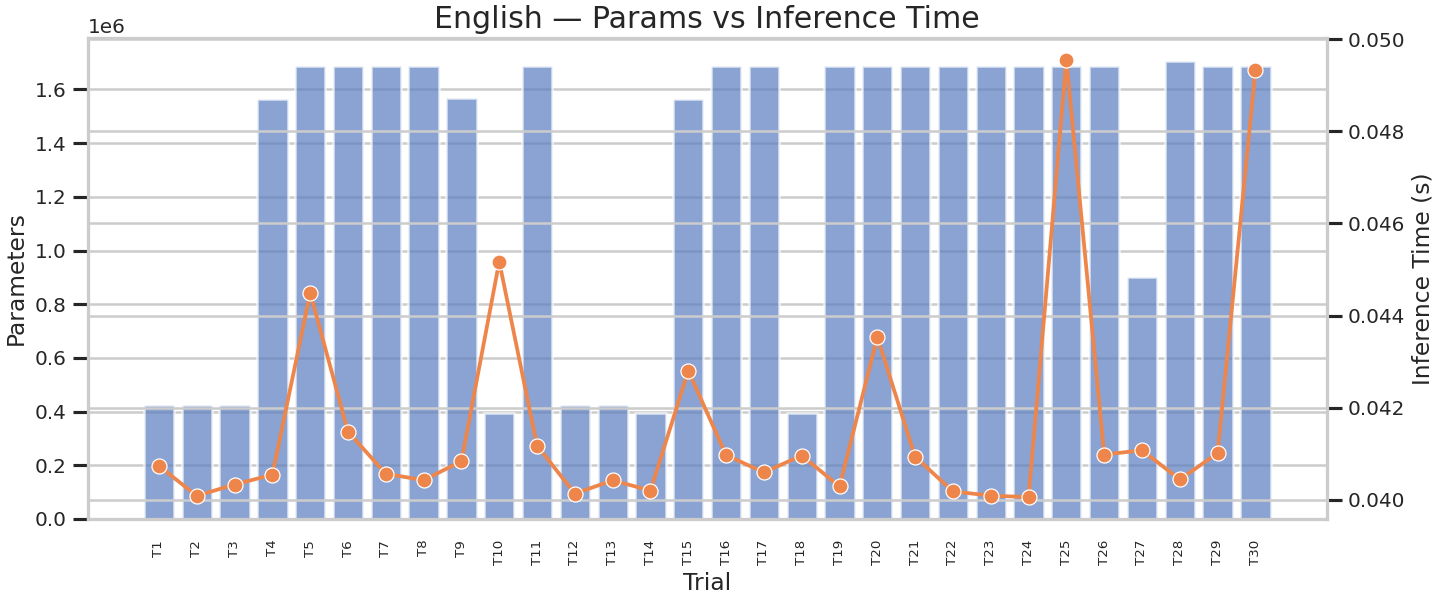}\hfill
    \includegraphics[width=.30\linewidth]{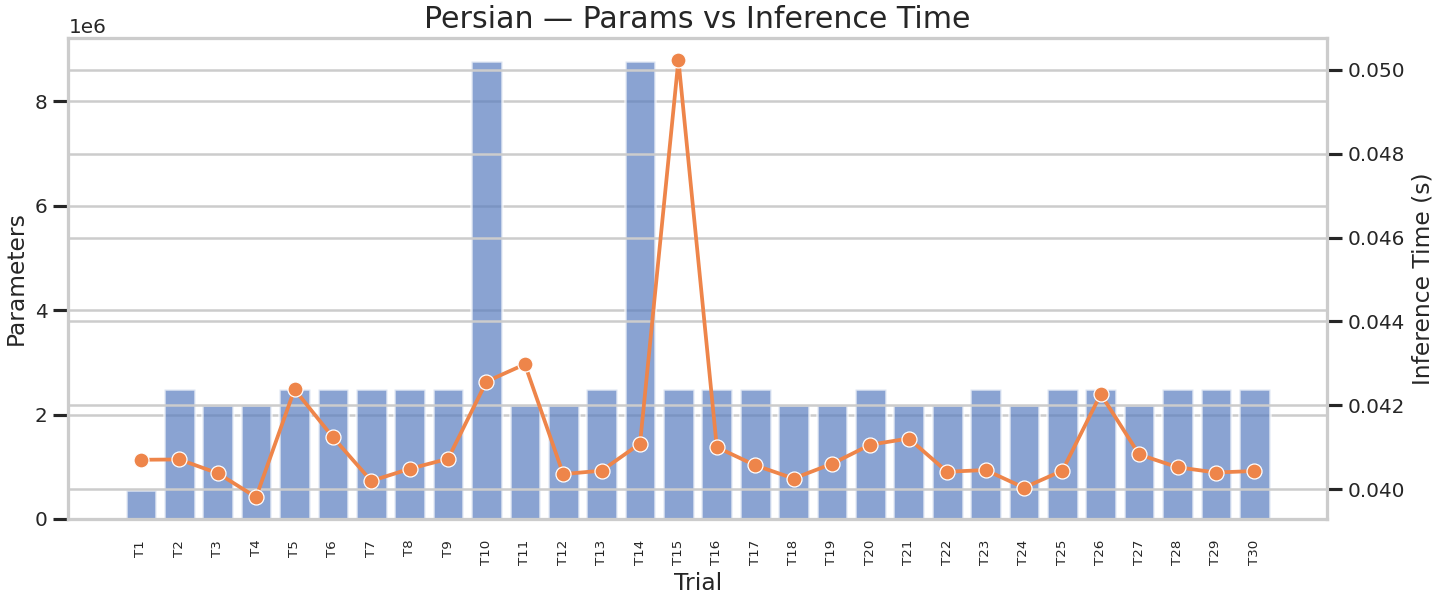}
    \caption{GPT 4o parameter count and per sample inference time across thirty trials for Arabic, English, and Persian. Bars show parameters and lines show mean latency. Latency remains near forty one milliseconds across a wide range of sizes.}
    \label{fig:gpt4o-paramlat}
\end{figure*}

\begin{figure*}[t]
    \centering
    \includegraphics[width=.30\linewidth]{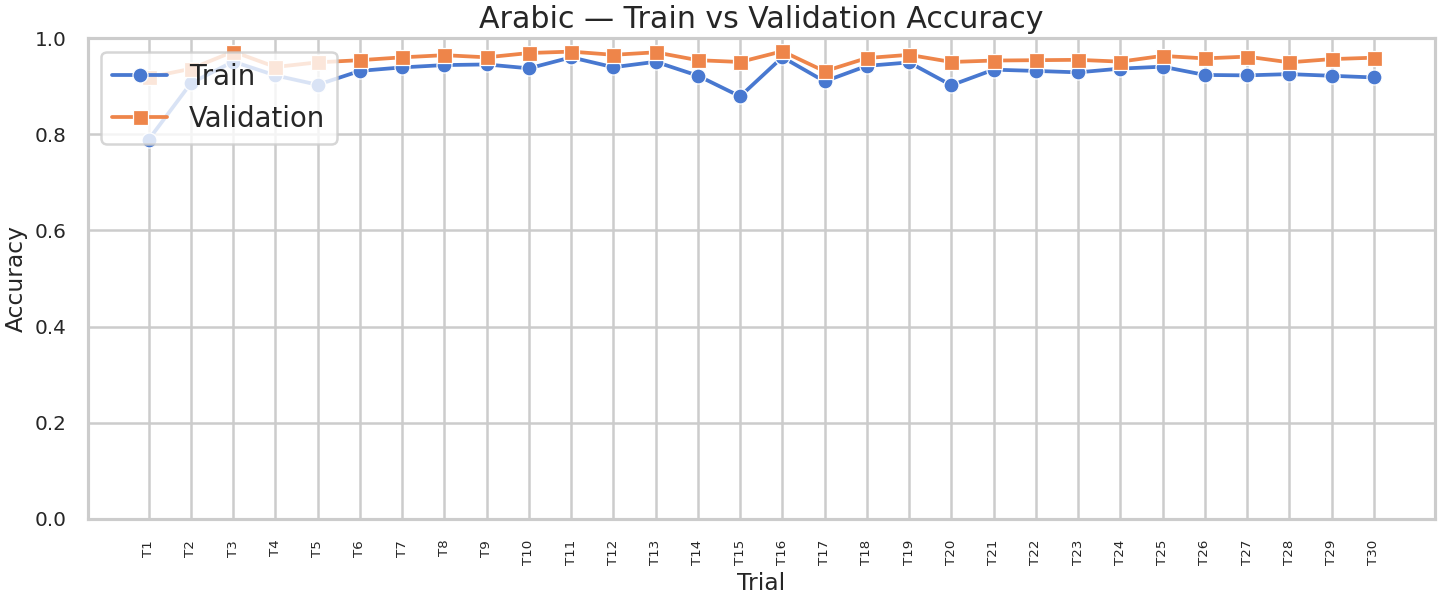}\hfill
    \includegraphics[width=.30\linewidth]{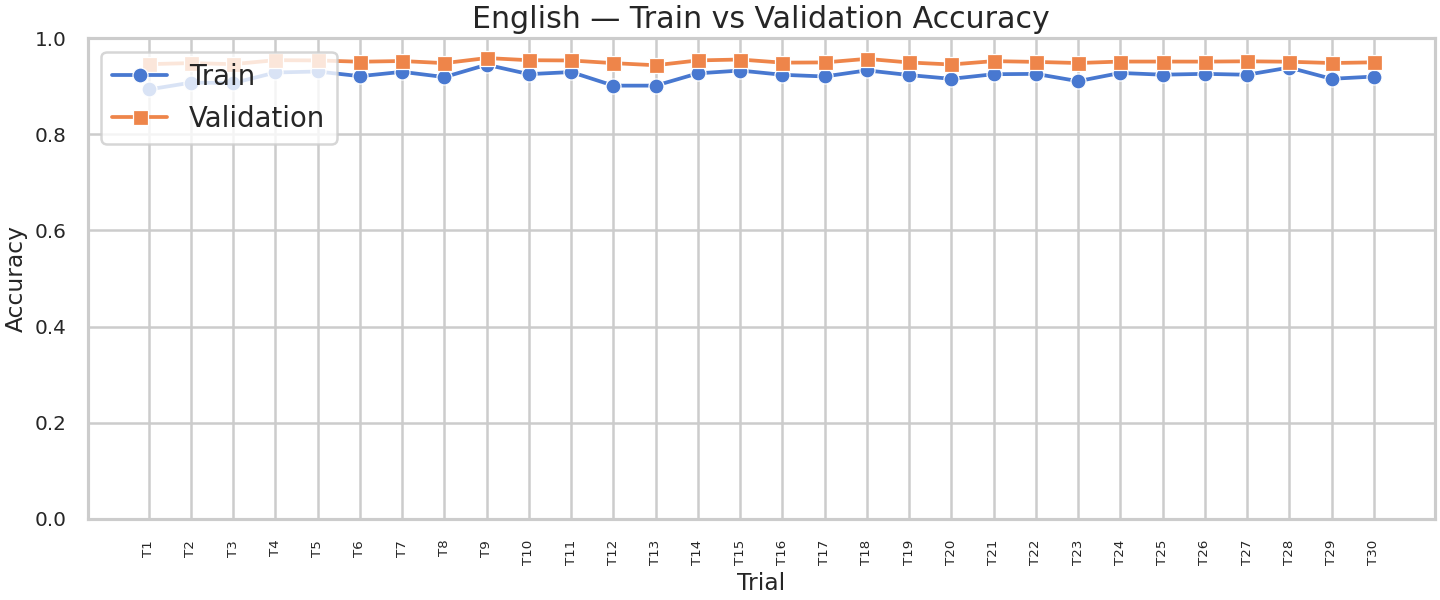}\hfill
    \includegraphics[width=.30\linewidth]{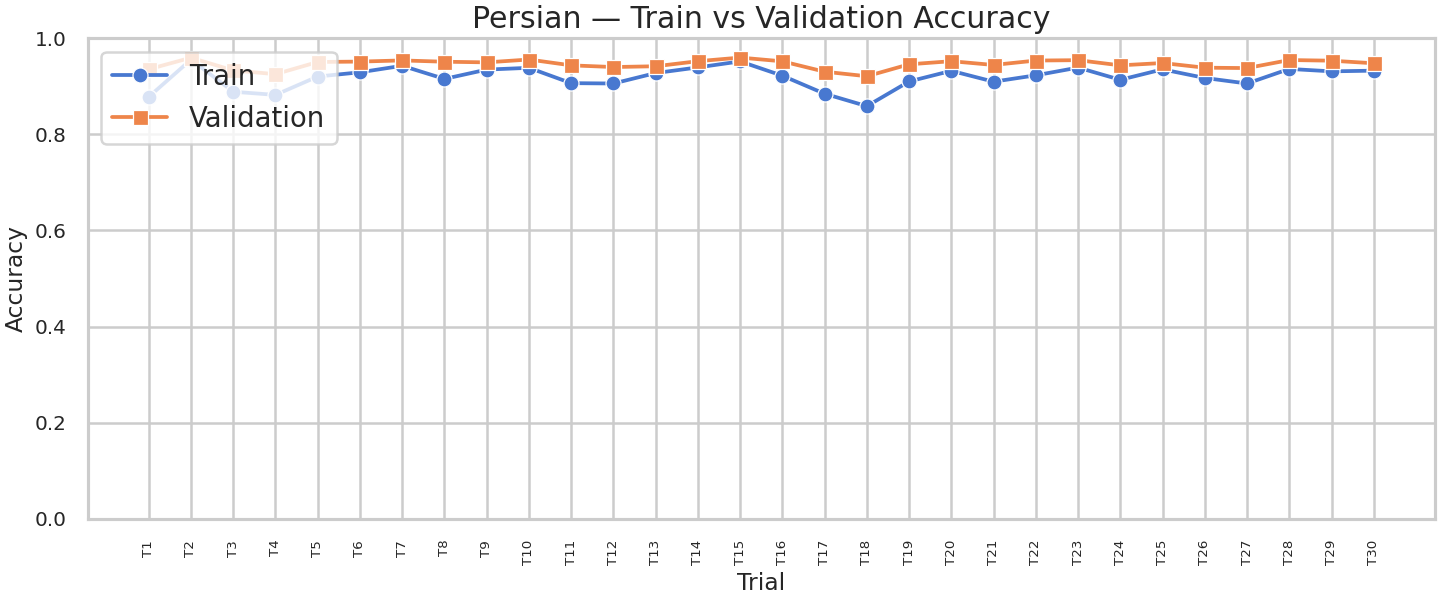}
    \caption{GPT 4o training and validation accuracy across thirty trials for Arabic, English, and Persian. Validation accuracy tracks training accuracy with small gaps, which indicates effective regularization and limited overfitting.}
    \label{fig:gpt4o-trainval}
\end{figure*}

\begin{figure}[t]
    \centering
    \includegraphics[width=0.55\columnwidth]{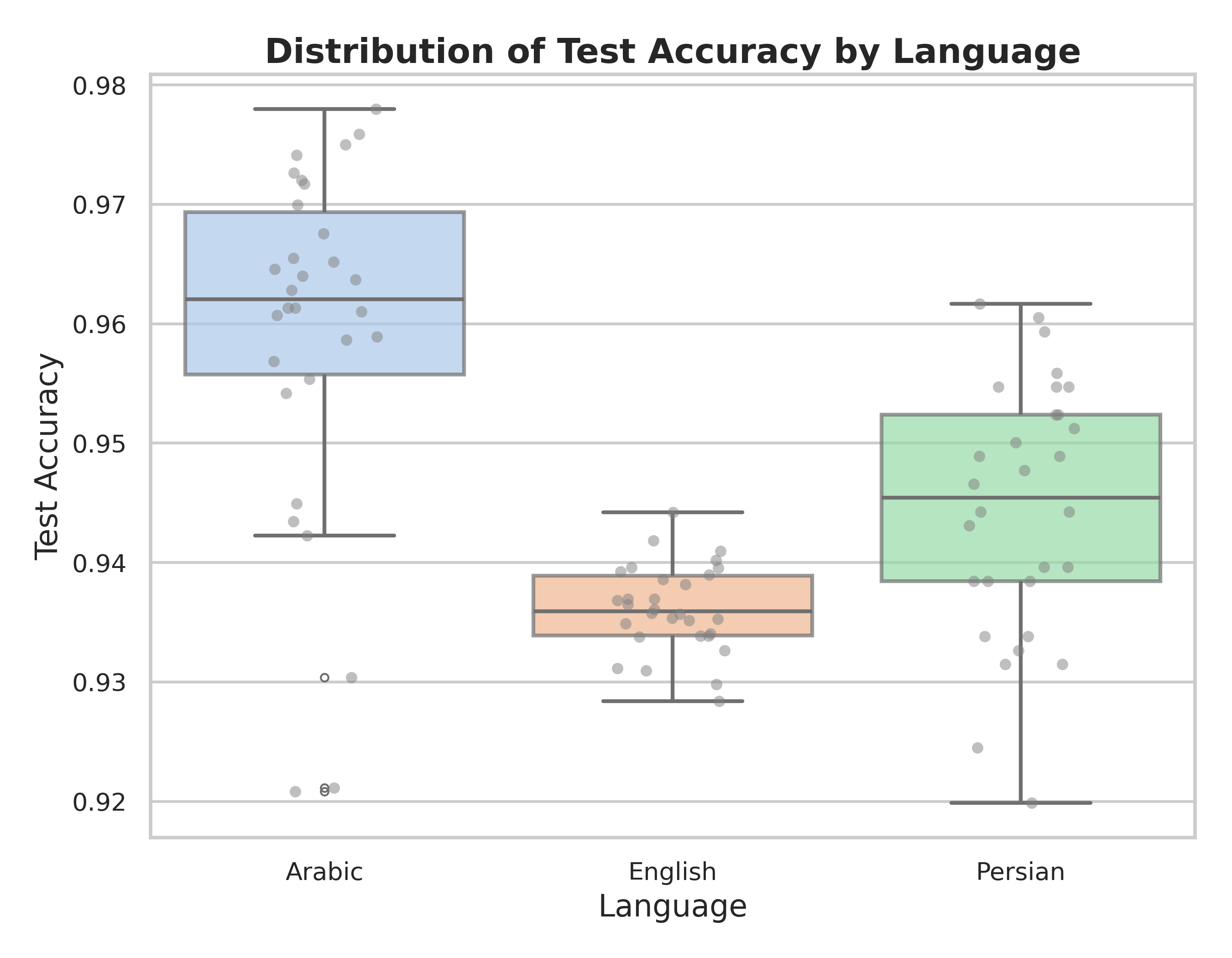}
    \caption{GPT 4o distribution of test accuracy across thirty trials per script for Arabic, English, and Persian. Arabic attains the highest median, English is slightly lower but very stable, and Persian is comparable.}
    \label{fig:gpt4o-boxplot}
\end{figure}

\begin{figure*}[t]
    \centering
    \includegraphics[width=.30\linewidth]{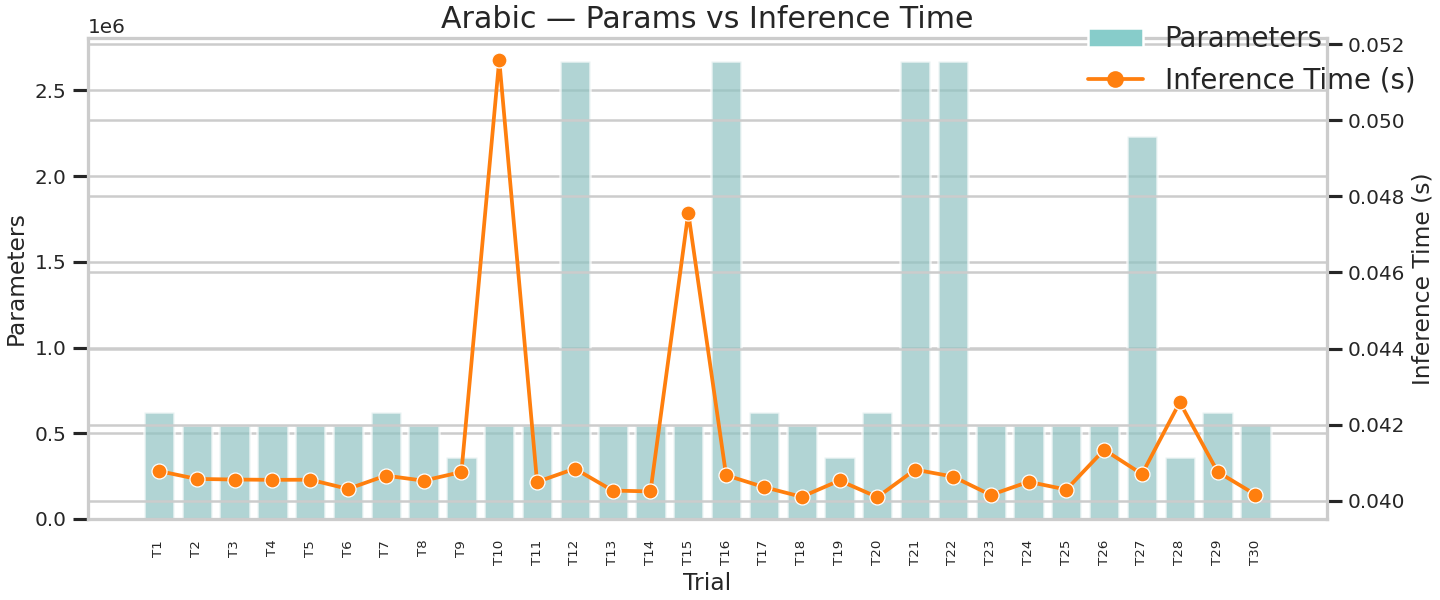}\hfill
    \includegraphics[width=.30\linewidth]{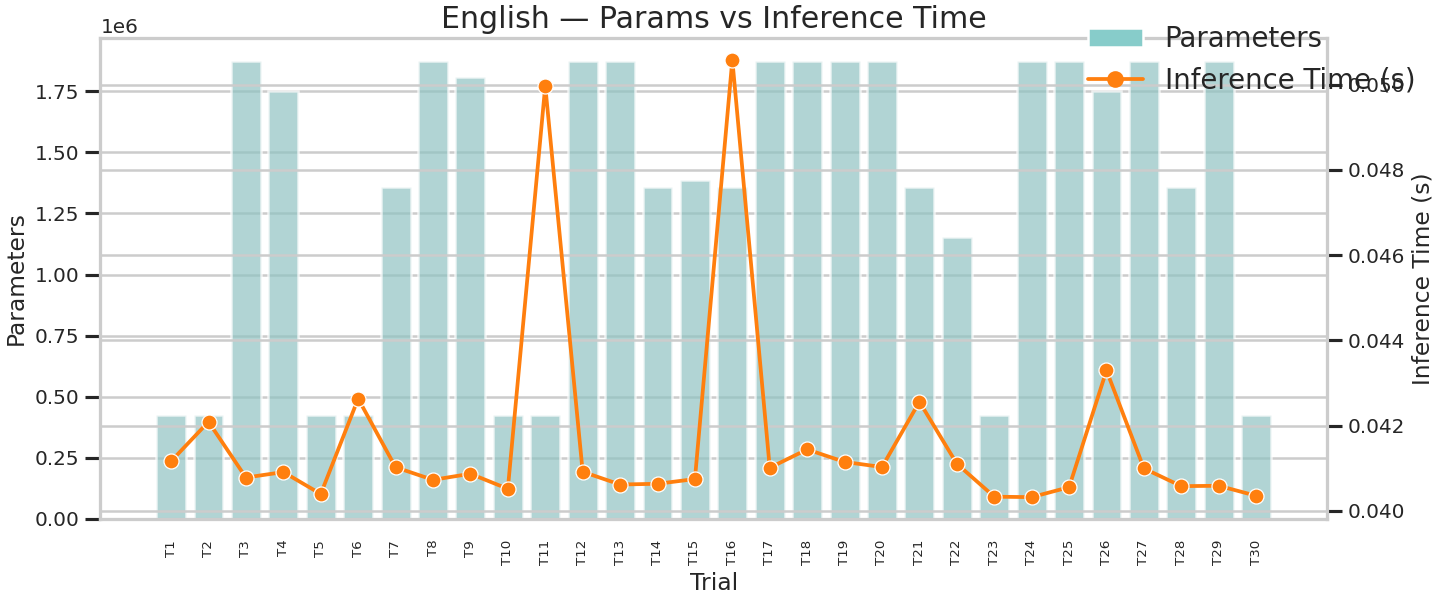}\hfill
    \includegraphics[width=.30\linewidth]{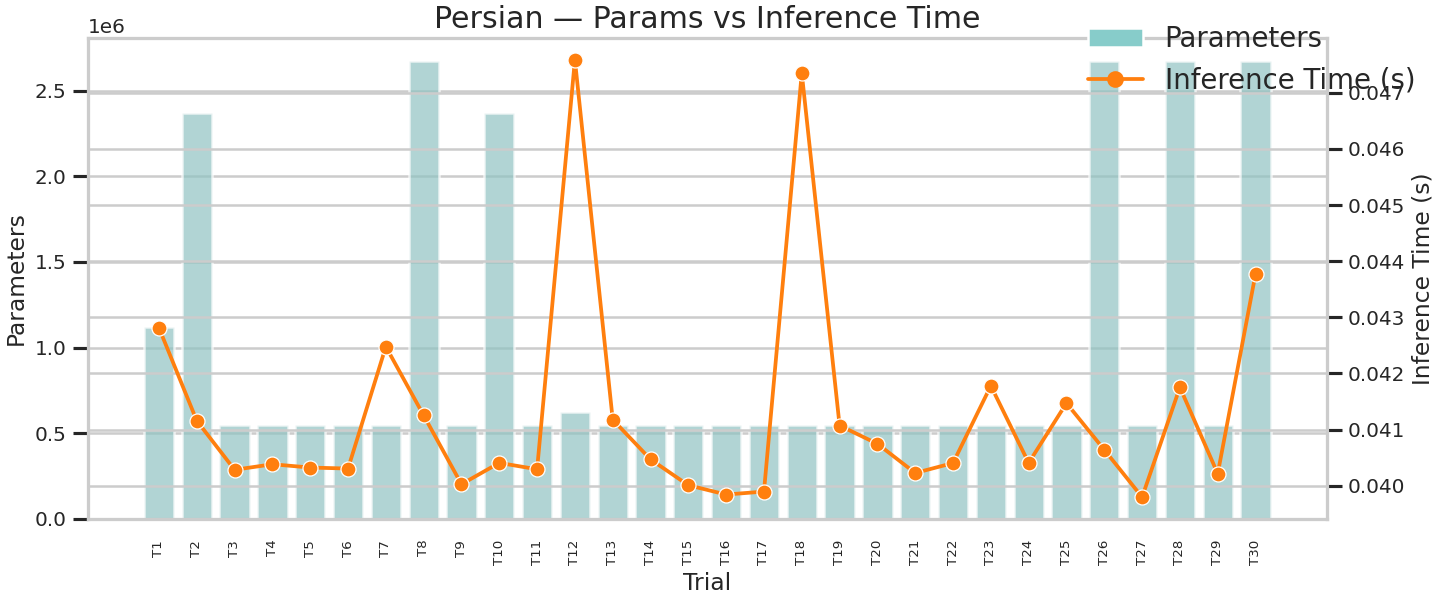}
    \caption{GPT 5 parameter count and per sample inference time across thirty trials for Arabic, English, and Persian. Bars indicate parameters and lines indicate mean latency. Latency remains near forty one milliseconds even when model size varies by several fold.}
    \label{fig:gpt5-paramlat}
\end{figure*}

\begin{figure*}[t]
    \centering
    \includegraphics[width=.32\linewidth]{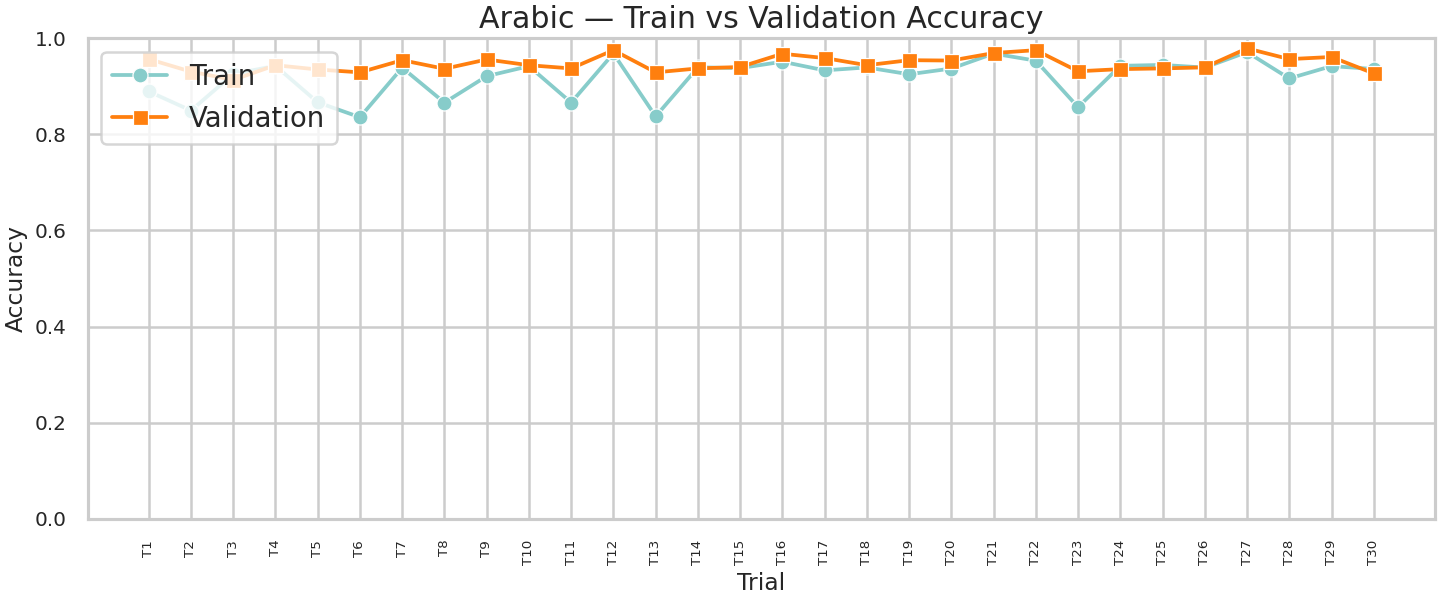}\hfill
    \includegraphics[width=.32\linewidth]{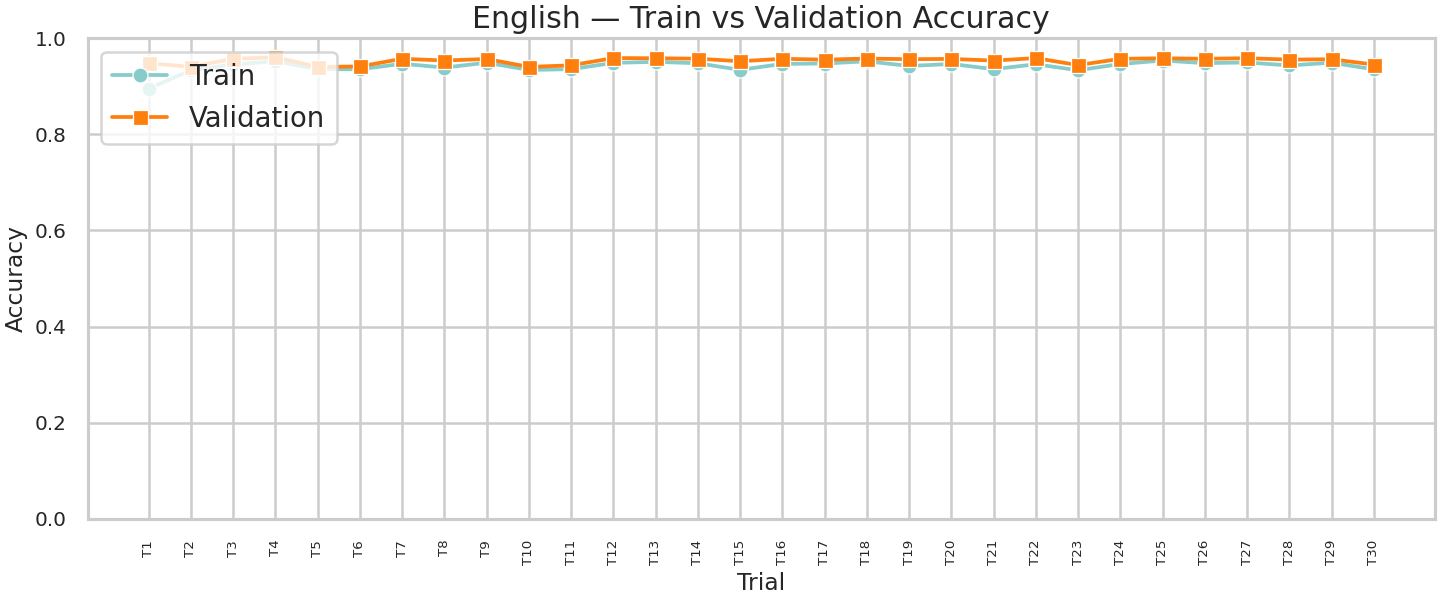}\hfill
    \includegraphics[width=.32\linewidth]{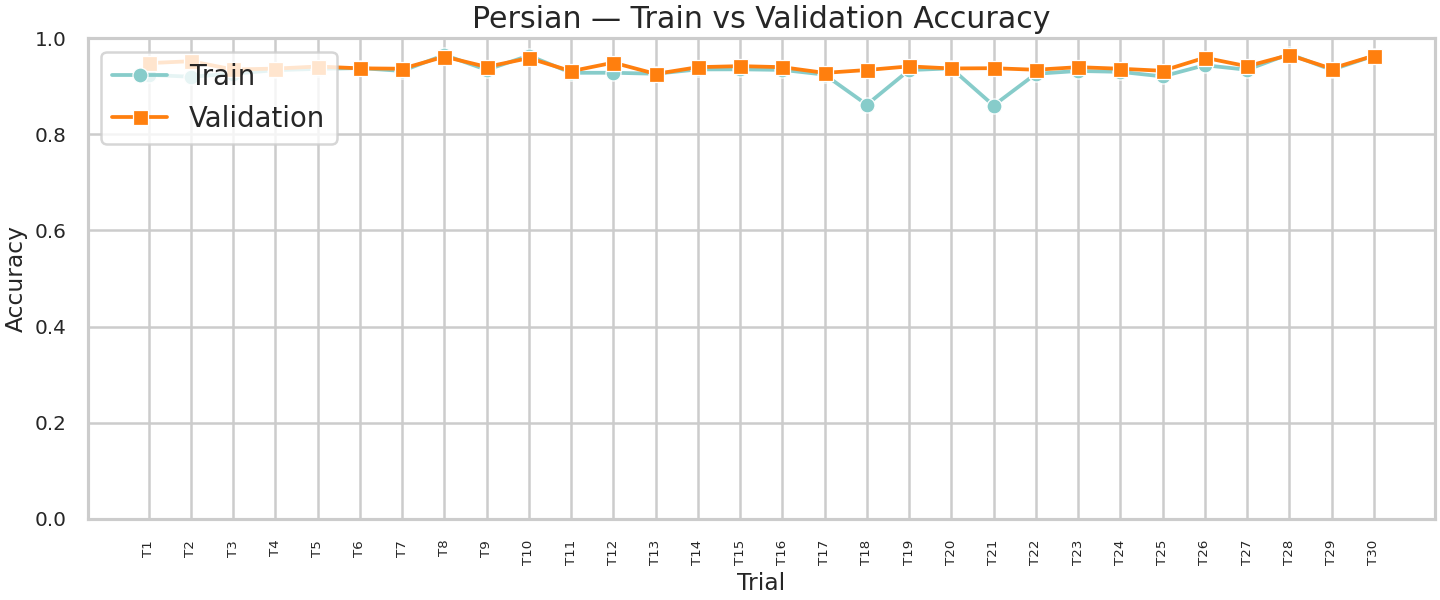}
    \caption{GPT 5 training and validation accuracy across thirty trials for Arabic, English, and Persian. Validation closely follows training with small gaps, which confirms strong generalization.}
    \label{fig:gpt5-trainval}
\end{figure*}

\begin{figure}[t]
    \centering
\includegraphics[width=0.4\columnwidth]{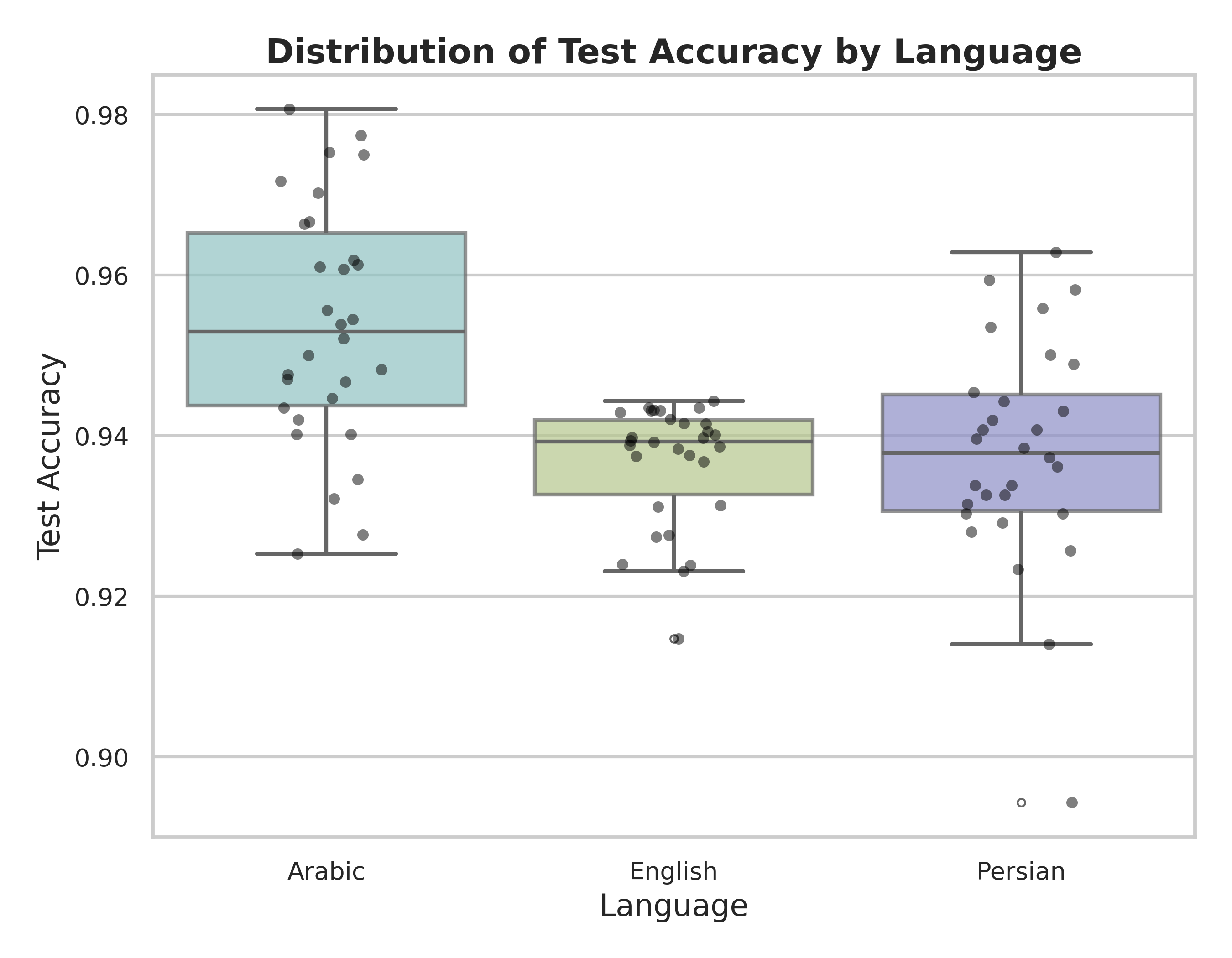}
    \caption{GPT 5 distribution of test accuracy across thirty trials per script for Arabic, English, and Persian. Arabic reaches the highest median with tight spread, English is slightly lower but stable, and Persian is competitive with a few low outliers.}
    \label{fig:gpt5-boxplot}
\end{figure}

\begin{table}[!t]
    \centering
    \caption{Aggregate performance across thirty independent trials per script.}
    \label{tab:agg-all}
    \captionsetup[sub]{font=footnotesize}

    \begin{subtable}[t]{\columnwidth}
        \caption{ChatGPT 5}
        \label{tab:agg-results_gpt5}
        \vspace{2pt}
        \resizebox{\columnwidth}{!}{%
        \begin{tabular}{lrrrrrr}
            \toprule
            Script & \multicolumn{1}{c}{Mean Acc} & \multicolumn{1}{c}{Std Dev} & \multicolumn{1}{c}{Best Trial} & \multicolumn{1}{c}{Best Acc} & \multicolumn{1}{c}{Mean Params (M)} & \multicolumn{1}{c}{Mean Latency (ms)}\\
            \midrule
            English & 0.937 & 0.008 & 12 & 0.944 & 1.35 & 41.7 \\
            Persian & 0.938 & 0.014 & 8  & 0.963 & 0.97 & 41.3 \\
            Arabic  & 0.954 & 0.015 & 27 & 0.981 & 0.88 & 41.2 \\
            \bottomrule
        \end{tabular}}
    \end{subtable}

    \vspace{4pt}

    \begin{subtable}[t]{\columnwidth}
        \caption{ChatGPT-4o}
        \label{tab:agg-results_gpt4}
        \vspace{2pt}
        \resizebox{\columnwidth}{!}{%
        \begin{tabular}{lrrrrrr}
            \toprule
            Script & \multicolumn{1}{c}{Mean Acc} & \multicolumn{1}{c}{Std Dev} & \multicolumn{1}{c}{Best Trial} & \multicolumn{1}{c}{Best Acc} & \multicolumn{1}{c}{Mean Params (M)} & \multicolumn{1}{c}{Mean Latency (ms)}\\
            \midrule
            English & 0.936 & 0.004 & 9  & 0.944 & 1.31 & 41.6 \\
            Persian & 0.944 & 0.011 & 16 & 0.962 & 2.74 & 41.2 \\
            Arabic  & 0.959 & 0.015 & 11 & 0.978 & 2.57 & 40.6 \\
            \bottomrule
        \end{tabular}}
    \end{subtable}

    \vspace{4pt}

    \begin{subtable}[t]{\columnwidth}
        \caption{Claude Sonnet 4}
        \label{tab:agg-resultsclaude}
        \vspace{2pt}
        \resizebox{\columnwidth}{!}{%
        \begin{tabular}{lrrrrrr}
            \toprule
            Script & \multicolumn{1}{c}{Mean Acc} & \multicolumn{1}{c}{Std Dev} & \multicolumn{1}{c}{Best Trial} & \multicolumn{1}{c}{Best Acc} & \multicolumn{1}{c}{Mean Params (M)} & \multicolumn{1}{c}{Mean Latency (ms)}\\
            \midrule
            English & 0.931 & 0.004 & 2  & 0.946 & 6.76 & 43.8 \\
            Persian & 0.946 & 0.007 & 1  & 0.964 & 9.28 & 42.3 \\
            Arabic  & 0.921 & 0.015 & 17 & 0.979 & 9.26 & 43.2 \\
            \bottomrule
        \end{tabular}}
    \end{subtable}
\end{table}

\begin{table}[!t]
    \centering
    \caption{Comparison of our AutoML-based LLM OCR framework and Cerescu \& Bumbu~\cite{b14}.}
    \label{tab:prior-compare}
    \vspace{2pt}
    \resizebox{\columnwidth}{!}{%
    \begin{tabular}{p{3cm}p{4.1cm}p{4.1cm}}
        \toprule
        & \textbf{Cerescu \& Bumbu (2024)} & \textbf{This Work} \\
        \midrule
        \textbf{LLM Model(s)}      & GPT-4o              & GPT-5, GPT-4o, Claude Sonnet 4 \\
        \textbf{LLM Role}          & Vision-based OCR    & Generative AutoML agent \\
        \textbf{Dataset(s)}        & Romanian Cyrillic   & Arabic, Persian, English \\
        \textbf{Preprocessing}     & Manual              & Fully automated \\
        \textbf{Cross-Lingual}     & No                  & Yes \\
        \textbf{Architecture Design} & Fixed             & Automated and dynamic \\
        \textbf{Loop Structure}    & One-pass evaluation & Iterative search and feedback \\
        \textbf{Accuracy Metric}   & Character-level     & Full-sequence accuracy \\
        \bottomrule
    \end{tabular}}
\end{table}
\vspace{-4pt}  

We evaluated the proposed closed loop neural architecture search pipeline over thirty fully independent trials for each script, Arabic, English, and Persian, for each language model designer. In every trial the assistant generated a new network topology and training schedule. We reported peak training, validation, and test accuracy, total parameters, and mean per sample latency averaged over one thousand forward passes on an NVIDIA GPU with batch size one and float32 precision. As summarized in Tables~\ref{tab:agg-results_gpt5}, \ref{tab:agg-results_gpt4}, and \ref{tab:agg-resultsclaude}, GPT 5 discovered the most compact architectures with mean parameter counts from 0.88 to 1.35 million and latencies from 41.2 to 41.7 ms, while reaching mean test accuracies of 0.954 for Arabic, 0.938 for Persian, and 0.937 for English, and a best trial of 0.981 on Arabic. GPT 4o closely matched or exceeded these accuracies, 0.959 for Arabic, 0.944 for Persian, and 0.936 for English, with slightly larger models between 1.31 and 2.74 million parameters and similar latency between 40.6 and 41.6 ms. Claude Sonnet 4 produced competitive models, especially for Persian with a mean of 0.946 and a best of 0.964, but with substantially larger sizes from 6.76 to 9.28 million parameters while maintaining real time latency from 42.3 to 43.8 ms. Variability across trials was low for all settings, with standard deviation between 0.004 and 0.015, and English typically showed the tightest spread. The training and validation curves for all trials show validation closely following training within about 0.5 to 2 percent, as seen in Figures~\ref{fig:gpt5-trainval}, \ref{fig:gpt4o-trainval}, and \ref{fig:claude-trainval}, which indicates robust generalization and effective early stopping. The parameter versus latency plots in Figures~\ref{fig:gpt5-paramlat}, \ref{fig:gpt4o-paramlat}, and \ref{fig:claude-paramlat} show that latency remains clustered near 40 to 44 ms despite multi fold changes in parameter count, which suggests that runtime cost is driven more by architectural depth and composition than by raw size. The accuracy distributions in Figures~\ref{fig:gpt5-boxplot}, \ref{fig:gpt4o-boxplot}, and \ref{fig:claude-boxplot} highlight cross lingual trends. Arabic attains the highest medians for GPT 5 and GPT 4o with tight variance, English is slightly lower but very stable, and Persian is competitive with a few low outliers and the strongest median for Claude Sonnet 4. Across ninety trials for each model, the pipeline consistently discovers accurate and parameter efficient architectures that meet real time constraints without manual tuning, which validates language model driven architecture search for multilingual handwritten optical character recognition.

\subsection*{A. Comparison with Prior Work}
A closely related study by Cerescu and Bumbu \cite{b14} compared GPT-4o’s vision-based OCR capabilities against traditional neural networks for a historical Romanian Cyrillic script, evaluating both models as direct recognition tools. Their work demonstrates the utility of LLMs for low-resource OCR, particularly at the character level, achieving up to 64\% accuracy with the GPT-4o API for single-character recognition. However, their approach treats LLMs solely as recognizers, without addressing model generation or adaptation. In contrast, our framework leverages GPT-5, GPT-4o and Claude Sonnet 4 as generative agents within an automated neural architecture search process, enabling not only recognition but also dynamic model design, script adaptation, and cross-lingual transfer across English, Persian, and Arabic datasets.

\subsection*{B. Comparison of LLMs}

A quantitative comparison of GPT 5, GPT 4o, and Claude Sonnet 4 shows clear differences in model quality and resource efficiency when used as automated neural architecture designers for multilingual handwritten text recognition. Summary statistics appear in Tables~\ref{tab:agg-results_gpt5}, \ref{tab:agg-results_gpt4}, and \ref{tab:agg-resultsclaude}.

\subsubsection*{1. Test Accuracy and Robustness}
GPT 4o attains the highest mean test accuracy for Arabic at 0.959, followed by GPT 5 at 0.954 and Claude Sonnet 4 at 0.921. Claude Sonnet 4 leads on Persian at 0.946, with GPT 4o at 0.944 and GPT 5 at 0.938. For English the means are 0.937 for GPT 5, 0.936 for GPT 4o, and 0.931 for Claude Sonnet 4. Best trial accuracies extend from 0.944 to 0.981 across the three models and scripts. GPT 5 reaches 0.981 on Arabic and 0.963 on Persian. GPT 4o reaches 0.978 on Arabic and 0.962 on Persian. Claude Sonnet 4 reaches 0.979 on Arabic and 0.964 on Persian and 0.946 on English. Variability is low in all cases, with standard deviation values between 0.004 and 0.015 across the three scripts and models. English shows the tightest variance at 0.004 for GPT 4o and Claude Sonnet 4 and 0.008 for GPT 5. Arabic shows 0.015 for all three models. These results indicate stable performance over thirty trials per script.

\subsubsection*{2. Model Size and Latency}
GPT 5 produces the most compact architectures with mean parameter counts of 1.35 million for English, 0.97 million for Persian, and 0.88 million for Arabic, and with mean latency from 41.2 to 41.7 ms. GPT 4o yields slightly larger models at 1.31 million for English, 2.74 million for Persian, and 2.57 million for Arabic, with latency from 40.6 to 41.6 ms. Claude Sonnet 4 generates substantially larger models at 6.76 million for English and about 9 million for Persian and Arabic, while maintaining real time inference from 42.3 to 43.8 ms. The small differences in latency despite large differences in parameter counts suggest that runtime cost is governed more by architectural depth and composition than by raw size, and all three approaches satisfy real time requirements.

\subsubsection*{3. Final Assessment}
GPT 5 offers the best accuracy to size profile, combining strong cross script accuracy with the smallest models and consistent real time latency. GPT 4o provides a balanced option with top mean accuracy in Arabic, competitive results in Persian and English, compact models, and very low latency. Claude Sonnet 4 achieves the strongest mean accuracy in Persian and competitive best trials across scripts, although it does so with a higher parameter budget.
\section{Conclusion and Future Work}
\label{sec:conclusion}
\balance
We presented a fully automated, closed-loop neural architecture search pipeline for multilingual handwritten character recognition, powered by GPT-5, GPT-4o, and Claude Sonnet 4. Experiments on Arabic, English, and Persian showed that the pipeline consistently discovers compact, accurate models without manual tuning or domain specific heuristics. Across trained models, it achieved strong generalization, efficient resource use, and inference times suitable for real-time deployment. Our analysis indicates that accuracy gains result from exploring depth and structural diversity rather than simply increasing parameter counts. Future work includes expanding to more languages with complex ligatures systems, integrating hardware-aware constraints, and extending to end-to-end sequence-to-sequence architectures for full line or paragraph recognition. These directions will further enhance the robustness, efficiency, and applicability of automated OCR model discovery.


\end{document}